\documentclass{article}

\usepackage{microtype}
\usepackage{graphicx}
\usepackage{subcaption}
\usepackage{booktabs} 
\usepackage{multirow}
\usepackage[accepted]{icml2020}

\icmltitlerunning{Disentanglement of Color and Shape Representations for Continual Learning}

\begin{document}

\twocolumn[
\icmltitle{Disentanglement of Color and Shape Representations for Continual Learning}




\begin{icmlauthorlist}
\icmlauthor{David Berga}{cvc}
\icmlauthor{Marc Masana}{cvc}
\icmlauthor{Joost Van de Weijer}{cvc}
\end{icmlauthorlist}

\icmlaffiliation{cvc}{Computer Vision Center, Bellaterra, Barcelona, Spain}

\icmlcorrespondingauthor{David Berga}{dberga@cvc.uab.es}

\icmlkeywords{Machine Learning, ICML}

\vskip 0.3in
]



\printAffiliationsAndNotice{} 
\begin{abstract}
We hypothesize that disentangled feature representations suffer less from catastrophic forgetting. As a case study we perform explicit disentanglement of color and shape, by adjusting the network architecture. We tested classification accuracy and forgetting in a task-incremental setting with Oxford-102 Flowers dataset. We combine our method with Elastic Weight Consolidation, Learning without Forgetting, Synaptic Intelligence and Memory Aware Synapses, and show that feature disentanglement positively impacts continual learning performance. 

\end{abstract}


\section{Introduction} \label{sec:introduction}
\label{sec:intro}
Convolutional Neural Networks have shown to increasingly achieve better performances in several recognition tasks over the past years \cite{Krizhevsky2012,LeCun2015,Guo2016}. In common image classification tasks, the network learns the whole dataset in a single training session. Thus, the network is only capable of doing inference on seen classes. If more classes would be learned without using a continual learning approach (i.e. finetuning on each task), the network would suffer of what is known as \emph{catastrophic forgetting} \cite{McCloskey1989, French1999, Kirkpatrick2017}. Catastrophic forgetting appears when neurons optimize their weights for a new task without taking into account previous knowledge; meaning previous classes performance is buried in favor of the new ones. Lifelong Learning and Continual Learning~(CL) propose a more realistic scenario where the learner continually adapts to a sequence of tasks while avoiding said catastrophic forgetting. One of the reasons of catastrophic forgetting could be the lack of adaptability from network parameters in the stability-plasticity trade-off from task to task, this interference can cause the network to entangle all trained representations.

\begin{figure}[ht]
    \begin{center}
    \includegraphics[width=\linewidth]{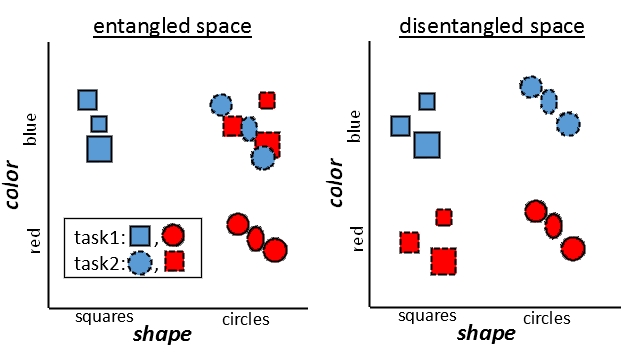}
    \caption{Illustration of feature disentanglement for a simple case of color "red/blue" and shape "square/circle". Left: Blue circles and red squares are entangled in the binding phase. Right: In a disentangled space, all shapes are separated along with colors.}
    \label{fig:drawn_color_shape}
    \end{center}
\end{figure}

Disentanglement of features, such as illumination, viewpoint object orientation or surface reflectance has been a long desired objective in computer vision~\cite{tappen2003recovering}. It is also believed to play an important role in the success of deep learning~\cite{bengio2009learning}. In this paper, we hypothesize that disentanglement of features also plays an important role in continual learning settings. The main idea is that disentangled features can better generalize to new tasks \cite{Bengio2013}. As an example 
(Fig.~\ref{fig:drawn_color_shape}), consider a system which should learn two tasks. The first task requires to distinguish between red circles and blue squares, while the second tasks requires to distinguish between blue circles and red squares. A network which would have learned a shape-color disentangled representation on the first task can easily adapt to the second task. However, a network which has learned an entangled representation (neurons firing for red-circles) might have more problems to generalize to other feature. While learning the second task, neurons specific to detect red-circles would instead detect blue-circles, consequently leading to catastrophic forgetting on the previous task.

Motivated by the biological evidence for separate processing of color and shape, we propose a two-branch network for image classification which fuses both at the end. This explicit color and shape disentanglement allows us to assess feature representation importance for continual learning and a way to fuse networks before binding.

\section{Related Work} \label{sec:relatedwork}

Latest reviews on CL \cite{Parisi2019,DeLange2019,Maltoni2019} 
focus on evaluating approaches by equally distributing classes from a dataset into multiple tasks. Those tasks are then learned in an incremental fashion by fitting the network parameters to each new group of classes. To avoid catastrophic forgetting, approaches regularize the model, store information or replay data \cite{Rebuffi2017,shin2017continual,lopez2017gradient}. Some of the first methods applied to neural networks, are focused on regularizing the weights or feature representations in order to keep those as close as possible to the older weights or representations while learning the new task. Learning without Forgetting (LwF) \cite{Li2016} adds a regularization term to the cross-entropy loss which tries to push the outputs of previous classes to be similar to the outputs of new tasks before learning the task at hand. Elastic Weight Consolidation (EWC) \cite{Kirkpatrick2017} calculates the Fisher Information for all weights, which is then used as an importance measure on the regularization loss. 
We consider these approaches in addition to Synaptic Intellingence (SI) \cite{Zenke2017} and Memory Aware Synapses (MAS)~\cite{Aljundi2018}. 

AlexNet \cite{Krizhevsky2012,Flachot2018} used 2 GPUs processing images in different groups of convolutions in parallel, sharing information in certain layers. By visualizing the filters in the first convolutional layer, the network showed sparse features that were distinct on the 2 network branches. These were similar to "gabor-like" filters for the gray branch and sinusoidal/concentric filters for the color branch, similar to receptive fields in V1 (see \cite{Krizhevsky2012}-Fig. 3 and \cite{Flachot2018}-Fig. 2). Rafegas et al. \cite{Rafegas2017,Rafegas2018,Rafegas2019} indexed selectivity of individual neurons to specific features in a VGG-M. By grouping these by color statistics (e.g. Hue), some appeared to be highly color selective and others low- or non color selective. Previous work~\cite{khan2012modulating} proposed an algorithm that processes shape and color separately (using SIFT features for shape features and color naming and pixel-wise hue descriptors for color features). Fusing these features showed improved accuracy in classification. From the aforementioned studies, we think a disentanglement procedure focused on separating color and shape feature computations in a network could be useful to acquire higher accuracy as well as preventing catastrophic interference, due to the specificity of neurons to each of these features. 

\section{Proposed Method} \label{sec:method}
In this paper, we propose disentangled architecture able to prevent catastrophic inference restricting specific feature representations. We process color and Shape features separately. We do so using a ResNet18-DS architecture (Fig.~\ref{fig:twobranch}) processing each feature representation separately in each branch (Fig.~\ref{fig:colorshape}). The \textit{objectives} are: a) Compute representations of Color and Shape separately. b) Learn disentangled representations for CL while retaining capacity. c) Provide an architectural mechanism able to be applied in combination with other CL algorithms (approach- and model-agnostic). 

\subsection{Independent Representations of Color and Shape} \label{sec:reps}

\begin{figure}[tb]
    \begin{center}
    \begin{subfigure}{.15\textwidth}
    \includegraphics[width=\linewidth,height=5.5cm]{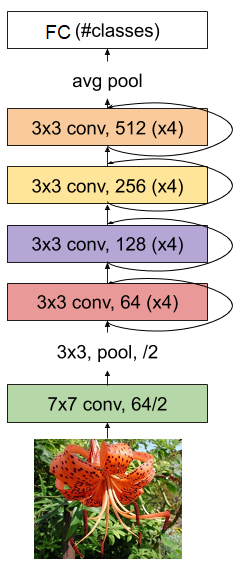}
    \caption*{ResNet18}
    \end{subfigure}
    \begin{subfigure}{.15\textwidth}
    \includegraphics[width=\linewidth,height=5.5cm]{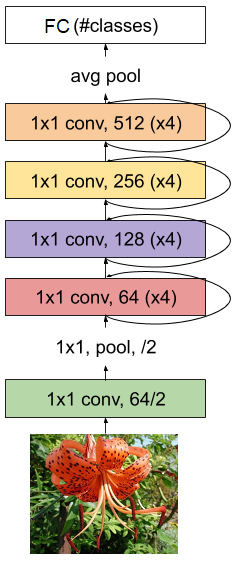}
    \caption*{ResNet18-Color}
    \end{subfigure}
    \begin{subfigure}{.15\textwidth}
    \includegraphics[width=\linewidth,height=5.5cm]{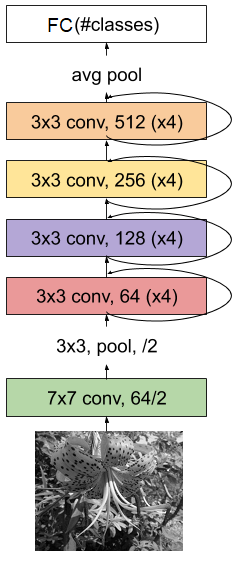}
    \caption*{ResNet18-Shape}
    \end{subfigure}
    \caption{Structure of the standard ResNet18 network, color and shape networks. Every layer (represented in a different color) sums~(+) its activation at every residual block.} \vspace{-1cm}
    \label{fig:colorshape}
    \end{center}
    \vspace{0.3em}
\end{figure}

We first discuss the networks we use for the computation of shape-only, and color-only features. Separation of these two features might better represent task-dependent combinations of color and shape (see Fig. \ref{fig:drawn_color_shape}). We consider two network parts: the \textit{Feature Extractor (FE)} and the \textit{Head / Classifier}. Here we use a ResNet18 as our base network, (Fig. \ref{fig:colorshape}-Left), but the idea could be applied to other models. The network feature extractor is composed of a convolutional+maxpooling layer, followed by four convolutional blocks with batch normalization, and an average pooling layer, and the classifier contains a fully-connected (FC). 

To uniquely process color (\textit{ResNet18-Color}), we changed all convolutional operations (in ResNet18 are conv7x7 in first layer and conv3x3 in all blocks) by a convolution of 1x1 kernels (conv1x1). This \textit{ResNet18-Color} network (Fig. \ref{fig:colorshape}-Mid) will learn each pixel independently from spatial computations, therefore it is only able to process color information. The absence of filters with spatial extend prevent it from being able to learn shape information. 

For the case of shape (\textit{ResNet18-Shape}), we transformed the RGB image to grayscale and used the original ResNet18 architecture. The \textit{ResNet18-Shape} network (Fig. \ref{fig:colorshape}-Right) is processing local information only using intensity information (thus, unable to process RGB chromaticities). 


\subsection{Fusing Color and Shape Representations in a Multi-Branch network (ResNet18-DS)}
Having established architecture which process color and shape separately in the previous section, we here propose an architecture to combine the two branches. The main requirements of our architecture are: \vspace{-0.5em}
\begin{itemize}
    \item \footnotesize \emph{Feature disentanglement}: The layers which are shared between the different tasks should only contain disentangled color and shape features. The entanglement should only happen in the task-specific head.
    \item \footnotesize \emph{Spatial binding}: the combination of color and shape should be entangled before any pooling operation.
\end{itemize}
\vspace{-0.5em}

\begin{figure}[tb]
    \centering
    \includegraphics[width=.65\linewidth,height=5.5cm]{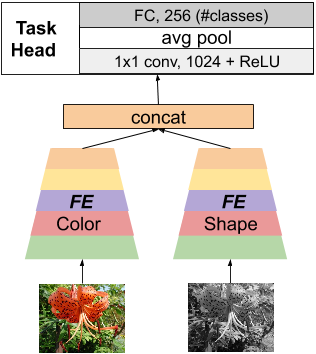}
    \vspace{-.5em}
    \caption{Disentangled network (ResNet18-DS) based on \textit{Late Fusion} of ResNet18-Color and ResNet18-Shape. The 3 layers on the Head (gray) are independent for each task.}
    \label{fig:twobranch}
\end{figure}

A system which would lack spatial binding is one where you would perform average pooling on the feature extractor of the color and shape branch and then combine the information. Such a system would know there are certain colors and shapes present, but could not say with certainty which color is connected to which shape.
In Fig.~\ref{fig:twobranch} the architecture which fulfills our design requirements is presented. The two-branch network forwards two versions of the input (color and gray) to the FE part of each branch (color and shape). Next the output of both branches is concatenated (the output of each branch is 7x7x512 and after concatenation the dimension is 7x7x1024).
Until here the color and shape information are disentangled and all layers are shared among the tasks. Then processing moves on to task-specific heads, in which the entanglement is performed. The task-specific head consists of four layers. First a 1x1 convolutional layer which maps the disentangled features, to entangled task specific features, followed by a ReLu. Then we a perform an average pooling operation to the output of the convolution and flatten to a feature vector. Finally, a Linear layer (FC) maps the entangled features to the number of classes of each particular task. A softmax operation is added after the FC before computing the gradient loss.

\begin{figure}[t]
\centering
\includegraphics[width=0.48\linewidth, height=2.85cm]{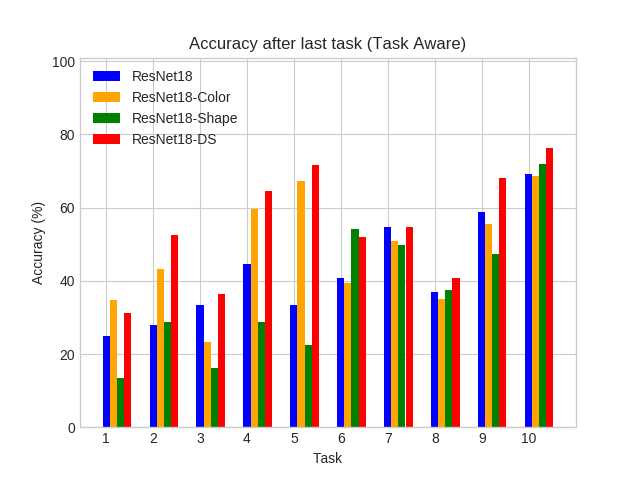}
\includegraphics[width=0.48\linewidth, height=2.85cm]{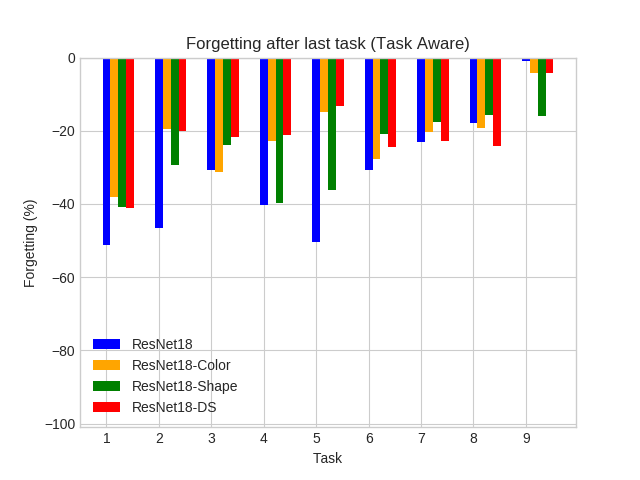}
\vspace{-0.4cm}
\includegraphics[width=0.48\linewidth, height=2.85cm]{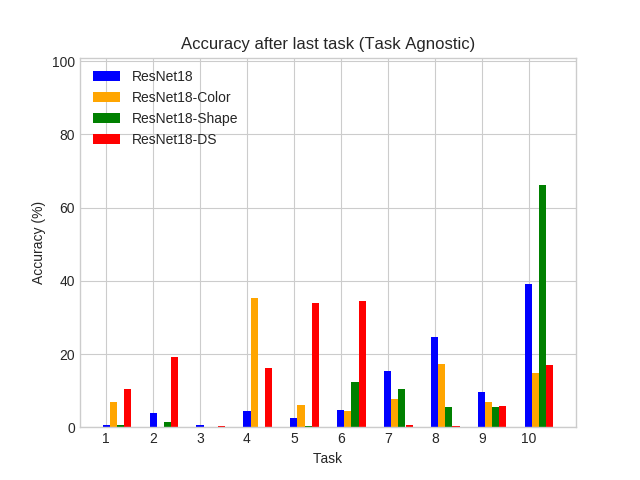}
\includegraphics[width=0.48\linewidth, height=2.85cm]{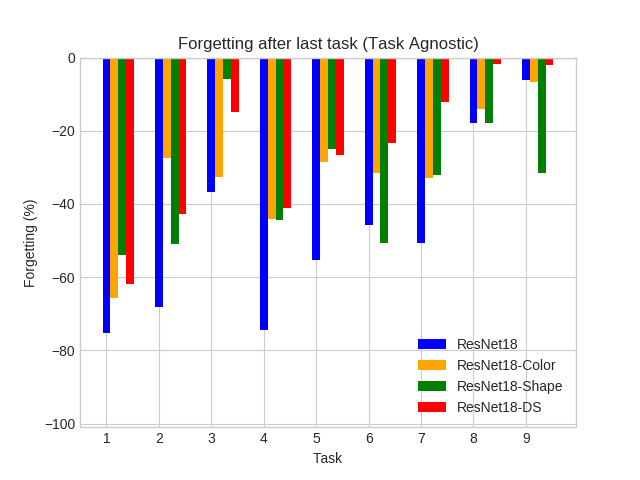}
\caption{Average classification results (Left: Accuracy, Right: Forgetting) per task in finetuning on Oxford-102 dataset.}
\label{fig:flowers_acc_taw}
\vspace{-0.4cm}
\end{figure}

\section{Experiments}\label{sec:experiments}

We performed a set of experiments with the standard ResNet18 as well as with the ResNet18-Color, ResNet18-Shape and our ResNet18-DS (Color+Shape) with the Oxford-102 dataset in an incremental setting of 10 tasks. To define this setting, we split the dataset (102 categories) into 10 tasks and processed the same network through each task split through the FE. We considered backpropagating the Head part for each task separately (task-aware) or concatenated in one unique Head (task-agnostic).  The capacity of the FE of ResNet-18 is 11.2M, while the capacity of Shape and Color branches FE are 11.2M and 1.5M respectively. The total capacity ResNet18-DS FE is 12.6M (about 112.5\% with respect the ResNet-18). For the each head (Linear Head and Task Head) there is a distinct capacity per task, given that we compared performance from original ResNet18 with same Task Head (ResNet18-H).

We processed distinct hyperparameters for each model to convergence (with learning rates 5e-2, 5e-3 and 5e-4 and batch size 32) for 200 epoch. We also applied a weight decay of 0.0002, momentum of 0.9 and we considered a patience of 15 (lowering the learning rates by a factor of 3 when loss does not improve after 15 epoch, until lr$<$1e-6). 

\begin{table*}[ht]
    \centering
    \small
    \begin{tabular}{cccccc}
    Task-Aware & ResNet18 & ResNet18-H* & ResNet18-Color & ResNet18-Shape & ResNet18-DS \\\midrule
    Finetune & 42.4 & 40.1 & 47.8 & 37.0 & 54.8 \\
    LwF & 58.3 & 55.8 & 53.8 & 50.9 & \textbf{62.8}  \\
    EWC & 46.4 & 45.4 & 52.9 & 41.3 & 55.5 \\
    MAS & 52.0 & 43.6 & 50.1 & 43.0 & 56.9 \\
    SI & 43.4 & 45.3 & 50.1 & 42.0 & 53.8 \\
    \end{tabular}
    \begin{tabular}{cccccc}
    Task-Agnostic & ResNet18 & ResNet18-H* & ResNet18-Color & ResNet18-Shape & ResNet18-DS \\\midrule
    Finetune & 10.6 & 10.2 & 10.0 & 10.2 & 13.8 \\
    LwF & 13.7 & 9.8 & 7.4 & 11.6 & 12.6 \\
    EWC & 10.3 & 11.3 & 12.4 & 9.1 & \textbf{17.0} \\
    MAS & 11.2 & 7.1 & 9.0 & 8.3 & 11.8 \\
    SI & 10.7 & 10.5 & 8.8 & 10.0 & 14.6 \\
    \end{tabular}
    \vspace{-0.3cm}
    \caption{Average accuracy classification results for Oxford-102 dataset in a incremental setting of 10 tasks. Mean Accuracy calculated from all tasks after the 10th task. \textbf{*} Same task-specific head (and head capacity) as ResNet18-DS (\hyperref[fig:twobranch]{Fig.\ref{fig:twobranch}}). \textbf{Bold} is TOP-1 accuracy.}
    \label{tab:flowers_10tasks_acc}
\end{table*}

\begin{table*}[ht]
    \centering
    \small
    \begin{tabular}{cccccc}
    Task-Aware & ResNet18 & ResNet18-H* & ResNet18-Color & ResNet18-Shape & ResNet18-DS \\\midrule
    Finetune & -29.2 & -29.8 & -19.8 & -24.0 & -19.3 \\
    LwF & -12.5 & -13.4 & -13.8 & -14.9 & -11.0  \\
    EWC & -22.6 & -24.9 & -14.3 & -20.2 & -17.1 \\
    MAS & -11.9 & -9.1 & -12.1 & \textbf{-4.6} & -9.4 \\
    SI & -26.1 & -25.3 & -14.7 & -19.4 & -18.0 \\
    \end{tabular}
    \begin{tabular}{cccccc}
    Task-Agnostic & ResNet18 & ResNet18-H* & ResNet18-Color & ResNet18-Shape & ResNet18-DS \\\midrule
    Finetune & -43.0 & -47.6 & -28.4 & -31.3 & -22.7 \\
    LwF & -31.6 & -35.6 & -24.1 & -38.7 & -23.5 \\
    EWC & -31.6 & -38.6 & -18.0 & -23.6 & -36.2 \\
    MAS & -9.5 & -6.9 & -8.3 & \textbf{-2.7} & -9.3 \\
    SI & -34.9 & -44.6 & -20.2 & -32.6 & -30.7 \\
    \end{tabular}
    \vspace{-0.3cm}
    \caption{Average forgetting for Oxford-102 dataset in a incremental setting of 10 tasks. Mean Forgetting calculated as drop of accuracy from all tasks after the 10th task. \textbf{*} Same task-specific head (and head capacity) as ResNet18-DS (\hyperref[fig:twobranch]{Fig.\ref{fig:twobranch}}).  \textbf{Bold} is lowest forgetting.}
    \label{tab:flowers_10tasks_forg}
    \vspace{-0.3cm}
\end{table*}

\begin{figure}[h!] 
\centering
\includegraphics[width=\linewidth,height=1.6cm]{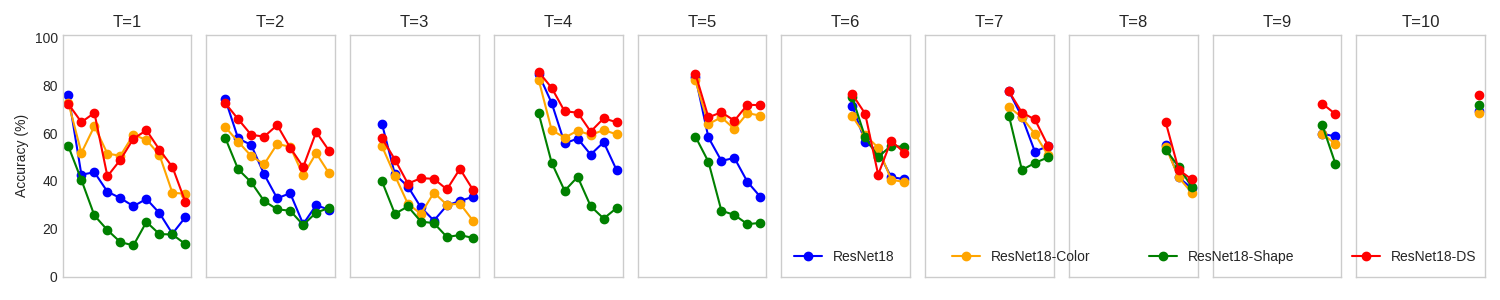}\\
\includegraphics[width=\linewidth,height=1.6cm]{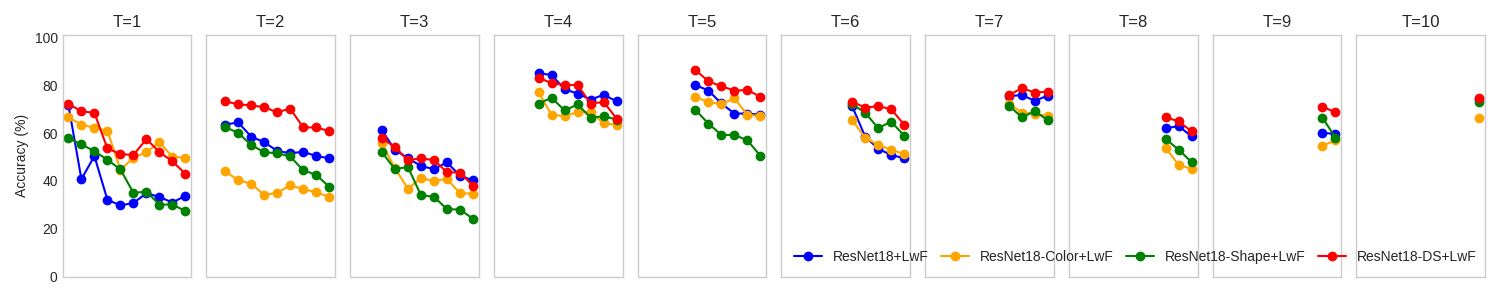}\\
\includegraphics[width=\linewidth,height=1.6cm]{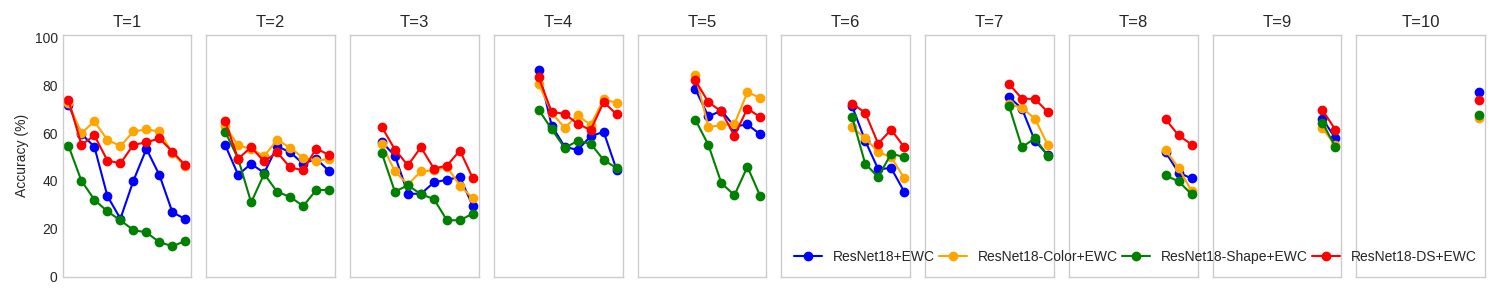}\\
\includegraphics[width=\linewidth,height=1.6cm]{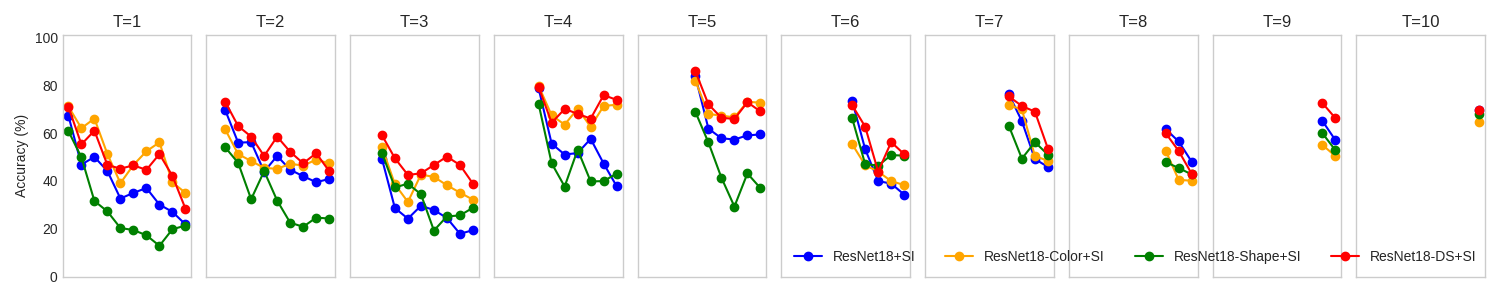}\\
\includegraphics[width=\linewidth,height=1.6cm]{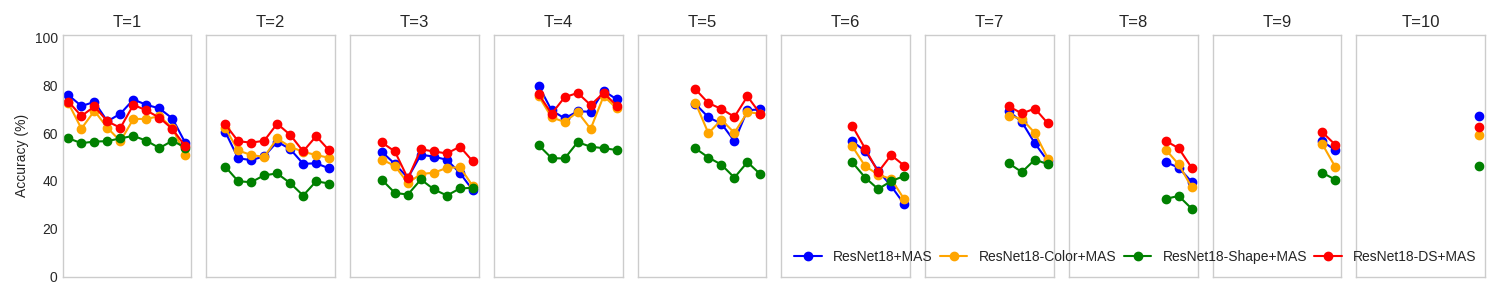}
\vspace{-0.6cm}
\caption{Task-sequence results for 10 tasks in Task-Aware setting for Finetune, LwF, EWC, SI and MAS.}
\vspace{-0.4cm}
\label{fig:flowers_tasks_acc_taw}
\end{figure}

In Table \ref{tab:flowers_10tasks_acc} we show results from the ablation results of each separate branch (ResNet18-Color and ResNet18-Shape). We have also processed the standard ResNet18 and our ResNet18-DS (Color+Shape) network with CL regularization algorithms (LwF, EWC, SI and MAS). Our model outperforms the standard state-of-the-art both in finetuning and with these two algorithms. Considering the classification results, our model presents lower forgetting with respect the standard ResNet, this means that our models is capable of disentangling activity of each task split at a feature level (the FE). In terms of accuracy per task our model performance is able to retain similar accuracy over tasks, whereas the standard ResNet18 accuracy lowers in both task-aware and task-agnostic settings. We showed in Figs. \ref{fig:flowers_acc_taw}-\ref{fig:flowers_tasks_acc_taw} that our model acquires overall higher accuracy and lower forgetting across all tasks. We believe this is due to the interference between features of the FE (specially at latter layers, which bind higher-level information). The standard ResNet shares all representations (e.g. color and shape) in a unique branch, which struggles on adapting weights for each new task representations. This interference prevented by our disentanglement procedure. We would like to point out that overall classification results for Task-Agnostic would be higher with exemplars. 

\section{Conclusion}

In this study we propose a novel architectural design that allows disentanglement of color and shape representations in a convolutional neural network. This method prevents catastrophic interference between these feature types, showing lower forgetting in comparison with standard networks. We show this strategy can be useful to be combined with other CL approaches (outperforming results based on standard architectures) and the potential to be used with distinct architectures and datasets (e.g. shapes, attributes). As such, we hope this paper inspires more research into the importance of disentangled representation for continual learning.

\vspace{-0.5em}
\section*{Acknowledgements}
We acknowledge the support from Huawei Kirin Solution. Marc Masana acknowledges 2019-FI\_B2-00189 grant from Generalitat de Catalunya.


\bibliography{egbib}

\end{document}